\documentclass[runningheads]{llncs}
\usepackage{amsmath} 
\usepackage{amssymb}
\usepackage{enumitem}
\usepackage{graphicx}
\usepackage{booktabs}
\usepackage{fancyvrb}
\usepackage{listings}     
\usepackage{algorithm}    
\usepackage{algpseudocode}
\usepackage{url}
\usepackage{xcolor}
\usepackage{listings}
\lstdefinelanguage{json}{
  morestring=[b]",
  morecomment=[l]{//},
  morekeywords={true,false,null},
  sensitive=false,
  alsoletter={:},
  moredelim=[l][\color{black}\bfseries]{"},
  moredelim=[s][\color{black}\bfseries]{:}{,},
}
\AtBeginDocument{%
  }

\newcommand{\nFDOs}{4{,}305}

\newcommand{\realConfAcc}{56.3}              
\newcommand{\nRealConf}{3{,}914}

\newcommand{\majorityAccClassC}{55.4}              

\newcommand{\clinvarRelease}{2026-05}
\newcommand{\clinvarURL}{ftp.ncbi.nlm.nih.gov/pub/clinvar/tab\_delimited/}
\newcommand{\nClinvarVariants}{8{,}978{,}110}      
\newcommand{\nClinvarSubmissions}{33{,}349}        
\newcommand{\cvFiltA}{1{,}851{,}125}               
\newcommand{\cvFiltB}{13{,}730}                    
\newcommand{\cvFiltC}{3{,}914}                     
\newcommand{\cvCatA}{42.2}                         
\newcommand{\cvCatB}{46.0}                         
\newcommand{\cvCatC}{11.8}                         

\newcommand{\nClassA}{1{,}744}                     
\newcommand{\nClassB}{1{,}918}                     
\newcommand{\nClassC}{65}                          
\newcommand{\nClassD}{187}                         

\newcommand{\ksBound}{0.522}                       
\newcommand{\tsPerturbPct}{20}                     
\newcommand{\tA}{29.8}                             
\newcommand{\tB}{10.7}                             
\newcommand{\tG}{3.6}                              

\newcommand{\nAdv}{16{,}438{,}800}                 
\newcommand{\nAdvRuns}{16{,}438{,}800}             
\newcommand{\advAccBound}{45.1}                    

\newcommand{\firstWinsAcc}{50.4}
\newcommand{\majorityAcc}{56.6}

\newcommand{\centLatRatio}{4.30}                   
\newcommand{\centThru}{6.75}                       
\newcommand{\autoPNF}{0.70}                        

\newcommand{\nRoundTrip}{45}                       

\newcommand{\rttMean}{144}                         
\newcommand{\rttStd}{55}

\newcommand{\implLOC}{8{,}742}
\newcommand{\anonRepo}{\url{https://github.com/zeyd31/aFDO_ISWC}}

\newcommand{\hwSetup}{\emph{Ubuntu 22.04.5 LTS VM with 8 vCPUs on an Intel Xeon Skylake-class processor, 15 GiB RAM, and 160 GB storage}}
\newcommand{\resMem}{4 GB}
\newcommand{\resCPU}{1.5 CPU-hours}

\newcommand{\trAa}{70.8} \newcommand{\trAb}{70.8} \newcommand{\trAc}{70.8}
\newcommand{\trAd}{70.8} \newcommand{\trAe}{70.8} \newcommand{\trAf}{70.8}
\newcommand{\trAg}{70.8}
\newcommand{\trBa}{58.4} \newcommand{\trBb}{58.4} \newcommand{\trBc}{58.5}
\newcommand{\trBd}{58.6} \newcommand{\trBe}{58.8} \newcommand{\trBf}{59.1}
\newcommand{\trBg}{58.9}
\newcommand{\trCa}{53.3} \newcommand{\trCb}{53.9} \newcommand{\trCc}{54.0}
\newcommand{\trCd}{54.0} \newcommand{\trCe}{53.9} \newcommand{\trCf}{54.0}
\newcommand{\trCg}{53.8}
\newcommand{\trDa}{52.9} \newcommand{\trDb}{53.5} \newcommand{\trDc}{53.5}
\newcommand{\trDd}{54.0} \newcommand{\trDe}{54.0} \newcommand{\trDf}{54.5}
\newcommand{\trDg}{54.5}
\newcommand{\trOa}{55.8} \newcommand{\trOb}{56.1} \newcommand{\trOc}{56.3}
\newcommand{\trOd}{56.3} \newcommand{\trOe}{56.4} \newcommand{\trOf}{56.5}
\newcommand{\trOg}{56.4}

\newcommand{\tsAa}{76.25}   
\newcommand{\tsAb}{21.0}    
\newcommand{\tsAc}{25.0}    

\newcommand{\tsBa}{22.5}
\newcommand{\tsBb}{21.0}
\newcommand{\tsBc}{23.25}

\newcommand{\tsCa}{23.25}
\newcommand{\tsCb}{21.0}
\newcommand{\tsCc}{27.5}
\newcommand{\tsDa}{23.0}
\newcommand{\tsDb}{22.5}
\newcommand{\tsDc}{28.75}
\newcommand{\tsEa}{20.75}
\newcommand{\tsEb}{23.25}
\newcommand{\tsEc}{31.25}

\newcommand{\accClassA}{58.6}     
\newcommand{\accClassB}{54.0}     
\newcommand{\accClassC}{70.8}                      
\newcommand{\accClassD}{54.0}                      

\newcommand{\consTimeout}{60}

\begin{document}

\title{Autonomous FAIR Digital Objects: From Passive Assertions to Active Knowledge}

\author{Zeyd Boukhers\inst{1,2}\thanks{Corresponding author: \email{zeyd.boukhers@fit.fraunhofer.de}} \and
Oya Deniz Beyan\inst{3,1} \and
Cong Yang\inst{4} \and
Christoph Lange\inst{1,5}}
\authorrunning{Z. Boukhers et al.}
%
\institute{
Fraunhofer Institute for Applied Information Technology FIT, Germany\\
\email{\{zeyd.boukhers, oya.deniz.beyan, christoph.lange-bever\}@fit.fraunhofer.de}
\and
University Hospital Cologne UKK, Cologne, Germany
\and
University of Cologne, Faculty of Medicine and University Hospital Cologne, Cologne, Germany
\and
School of Future Science and Engineering, Soochow University, Suzhou, China\\
\email{cong.yang@suda.edu.cn}
\and
RWTH Aachen University, Aachen, Germany
}

\titlerunning{Autonomous FAIR Digital Objects}

\maketitle

\begin{abstract}
Scientific knowledge on the Web is published as passive assertions and cannot decide when to validate evidence, reconcile contradictions, or update confidence as findings accumulate. Curation depends on centralised middleware and institutional continuity, but when registries close, active stewardship stops even when data remain online. We advance the concept of \emph{Autonomous FAIR Digital Objects} (aFDOs) from an abstract idea to an operational model, to offer a route from passive scientific publication toward accountable, standards-aligned automation that can outlive its publishing institutions. aFDO augments FDOs with three capabilities anchored in Semantic Web standards, namely 1) a policy layer over RDF-star aligned with PROV-O, SHACL, and ODRL for portable condition-action rules, 2) an announcement layer over ActivityStreams 2.0 that bounds per-announcement evaluation cost, and 3) an agreement layer that resolves multi-source contradictions through reputation and confidence weighted agreement under a bounded adversarial model. We provide a formal definition that distinguishes policy specifications, event handlers, and communication interfaces.
We evaluate an open reference implementation on \nFDOs{} FDOs grounded in rare-disease ontologies, namely ClinVar, HPO, and Orphanet, combined with controlled synthetic observations. The consensus mechanism resolves \realConfAcc{}\% of \nRealConf{} naturally occurring ClinVar conflicts where multiple submitters disagree and an expert panel has subsequently adjudicated. Under Sybil, collusion, and poisoning attacks, the mechanism degrades gracefully within its design Byzantine-tolerance bound ($f < n/5$), and fails as predicted beyond that bound.

\end{abstract}





\section{Introduction}
\label{sec:introduction}

The Hepatitis Delta International Network registry documented 2{,}007 patients before closing in 2022\footnote{\url{https://www.german-liverfoundation.com/hdin/}}. The Rory Morrison Waldenström's Macroglobulinaemia registry tracked 1{,}305 patients across 22 UK centres before ceasing submissions in 2025\footnote{\url{https://wm.e-dendrite.com/}}. The data remain online, indexed by persistent identifiers and described by rich metadata, but the networks can no longer validate new observations, reconcile conflicting assessments, or update confidence as evidence accumulates, meaning the artefacts survive, but active curation does not.

These examples are not only failures of institutional sustainability, but rather expose a broader limitation in how machine-readable scientific knowledge is published. FAIR and Semantic Web approaches make knowledge artefacts durable, identifiable, richly described, and reusable after the organisations that created them have changed or disappeared. Nanopublications, for example, package minimal scientific assertions with provenance~\cite{groth2010anatomy,kuhn2016nanopublications}, while FAIR Digital Objects bind content to persistent identifiers, types, metadata, and declared operations~\cite{wittenburg2022fair,lannom2022digital}. Both help formalise \emph{what} a knowledge artefact is and \emph{what can be done} with it, but neither specifies \emph{when} operations should be invoked, \emph{how} contradictions should be resolved across sources, or \emph{under what conditions} a claim's confidence should change. Those decisions remain in external middleware or institutional workflows, and when that layer stops, curation stops with it.
This raises three practical challenges. The first is \emph{continuous validation}, where new evidence arrives asynchronously and may contradict prior claims, but manual triage does not scale, and opaque automation erodes trust. The second is \emph{discovery under growth}, where relevant artefacts must be found in expanding corpora without exhaustive scans. The third is \emph{cross-institution coordination}, where different institutions may reach different conclusions about the same claim, so any automated approach must be able to compare evidence, weigh trust, and record how a shared interpretation was reached. These problems have been studied separately in active databases, publish-subscribe systems, and Byzantine consensus, but their integration with provenance-bearing FAIR artefacts remains underdeveloped.

This paper builds on the existing concept of \emph{Autonomous FAIR Digital Objects} (aFDOs)~\cite{lannom2022digital} and develops it into an operational model for active, auditable curation. The model extends FDOs with three capabilities, namely a \emph{policy layer} for condition-action rules and provenance-aware audit trails, an \emph{announcement layer} for routing relevant events between subscribed peers, and an \emph{agreement layer} for resolving conflicting claims across sources while preserving the evidence, trust weights, and policy versions used in the decision.
We are explicit about what decentralisation means here. aFDOs are \emph{logically decentralised} objects on a shared discovery substrate, where the registries play a coordinating role similar to that of DNS. Validation, agreement, and policy execution occur at the object layer without centralised decision orchestration. We discuss registry-free deployment through DHT-based discovery as future work, but do not claim to remove the shared discovery substrate in this paper.
The contributions of the paper are summarised as follows:

\begin{enumerate}[leftmargin=*,itemsep=2pt]
    \item A formal model of aFDOs that distinguishes policy specifications, event handlers, and communication interfaces, with each component mapped to W3C vocabularies, namely RDF-star, PROV-O, SHACL, ODRL, and ActivityStreams~2.0.
    \item A reference implementation integrating policy evaluation, subscription-based announcement, and reputation and confidence weighted agreement over nanopublication-style RDF artefacts, released as an open, reproducible artefact.
    \item An evaluation on \nFDOs{} FDOs grounded in rare-disease ontologies, namely ClinVar, HPO, and Orphanet, combined with controlled synthetic observations, reporting \realConfAcc{}\% accurate resolution of \nRealConf{} naturally occurring ClinVar conflicts where multiple submitters disagree and an expert panel has subsequently adjudicated, and graceful degradation under Sybil, collusion, and evidence-poisoning attacks within the design Byzantine tolerance ($f < n/5$).
    \item A discussion of standards alignment, deployment trade-offs, and limitations, including registry centralisation, trust bootstrapping, and economic sustainability.
\end{enumerate}

Section~\ref{sec:background} reviews nanopublications, FDOs, and the Semantic Web foundations the model builds on, and positions the work against autonomous agents, active databases, and decentralised RDF. Section~\ref{sec:afdo} formalises aFDOs and presents the architecture. Section~\ref{sec:evaluation} describes the reference implementation and reports the evaluation. Section~\ref{sec:conclusion} discusses limitations and future directions.

\section{Background and Related Work}
\label{sec:background}
This section reviews the foundations of aFDOs, namely nanopublications, FAIR Digital Objects, and the relevant Semantic Web standards. It then discusses related work on autonomous agents, active databases, decentralised RDF, and consensus mechanisms.

\subsection{Nanopublications and FAIR Digital Objects}

Nanopublications represent minimal scientific assertions as RDF graphs~\cite{groth2010anatomy}. Each nanopublication consists of three named graphs, $\langle G_{\text{assertion}}, G_{\text{provenance}}, G_{\text{pubinfo}}\rangle$, capturing respectively the scientific claim (e.g., a gene--disease association), how it was derived (methods, sources, timestamps), and publication metadata (authors, licences, etc.). Public repositories now contain millions of nanopublications spanning biomedicine, environmental, and social science~\cite{kuhn2016nanopublications}, and decentralised nanopublication services support direct publication and federated querying~\cite{kuhn2021nanobench}. The repositories nonetheless remain archives, in the sense that they record claims with rich provenance but do not validate assertions against new evidence, reconcile contradictions, or update confidence as findings accumulate.

FAIR Digital Objects (FDOs)~\cite{wittenburg2022fair,jacobsen2020fair,zoubia2024fdo} bind content to machine-actionable capabilities. An FDO is a quadruple $\langle \text{PID}, \mathcal{T}, \mathcal{O}, \mathcal{M}\rangle$, where the PID is a persistent identifier, such as Handle or DOI~\cite{lannom2022digital}, $\mathcal{T}$ specifies the type, $\mathcal{O}$ is the set of supported operations, and $\mathcal{M}$ is the respective metadata. Operations are invoked via the Digital Object Interface Protocol (DOIP), which provides a standard interface for interacting with digital objects. Clients dereference the PID, retrieve the FDO record, and invoke declared operations such as \texttt{create}, \texttt{retrieve}, \texttt{update}, or domain-specific extensions. Recent work compares FDOs with Linked Data~\cite{soilandreyes2024fdovsld,schultes2022comparative} and demonstrates FDO deployment in publishing workflows and as nanopublication extensions~\cite{schultes2024fdopublishers,magagna2024nanopubsasfdo}.

\subsection{From Passive Artefacts to Active Knowledge}

FDOs declare what \emph{can} be done with an artefact but not \emph{when}, \emph{how}, or under \emph{what} conditions an operation should be invoked. A patient observation FDO may declare \texttt{validate()}, but the decision of when to invoke it (a confidence threshold, elapsed time, or contradictory evidence) lives in client code or external middleware. A variant interpretation FDO may declare \texttt{reconcile()}, but conflict identification and orchestration across institutions falls outside the FDO itself. This is not a limitation of the FDO model, which provides a suitable primitive for FAIR publication. The issue is the placement of decision logic. When invocation conditions, conflict resolution, and trust updates live outside the artefact, the registry closures discussed in Section~\ref{sec:introduction} break the active curation while the data remain accessible. Table~\ref{tab:capabilities} summarises the resulting capability gap.

\begin{table}[t]
\centering
\caption{Comparison of capabilities in nanopublications, FAIR Digital Objects, and aFDOs.}
\label{tab:capabilities}
\small
\begin{tabular}{lccc}
\toprule
\textbf{Capability} & \textbf{Nanopublications} & \textbf{FDOs} & \textbf{aFDOs} \\
\midrule
Persistent identification & \checkmark & \checkmark & \checkmark \\
Provenance or rich metadata & \checkmark & \checkmark & \checkmark \\
Declared operations & $\times$ & \checkmark & \checkmark \\
\addlinespace
\multicolumn{4}{l}{\emph{Embedded decision logic}} \\
Invocation conditions & $\times$ & $\times$ & \checkmark \\
Multi-source agreement & $\times$ & $\times$ & \checkmark \\
Evidence-driven trust updates & $\times$ & $\times$ & \checkmark \\
\bottomrule
\end{tabular}
\end{table}

\subsection{Related Work}

\textbf{Autonomous agents and active systems.}
Autonomous agents and multi-agent systems provide models for coordination, reasoning, and distributed decision making~\cite{wooldridge2009introduction,bellifemine2007developing,bordini2007programming}. Hypermedia-based MAS extend these ideas to Web-scale agent ecosystems~\cite{ciortea2019hypermedia}, and Semantic Web agents use RDF and OWL for representation and reasoning~\cite{hendler2001semantic}. Active databases and event-driven systems support event-condition-action rules and publish-subscribe communication~\cite{paton1999active,michelson2006event}. In these approaches, the rules are usually placed in an agent, database, broker, or application service. In contrast, aFDOs represent policy conditions, provenance, and interaction rules with the FAIR artefact itself, while execution is still performed by software components.

\textbf{Decentralised RDF and knowledge graphs.}
Knowledge graph surveys mainly address representation, reasoning, and querying~\cite{hogan2021knowledgegraphs,chaudhri2022knowledgegraphs}. Decentralised RDF systems study peer-to-peer storage, indexing, and query processing~\cite{Aebeloe2019-ESWC,Aebeloe2019-ISWC,Cai2004-WWW,Zhou2025-PVLDB}. Other work supports shared updates with provenance or temporal access to RDF datasets~\cite{Aebeloe2021-WWW,Pelgrin2021-SWJ}. Blockchain-based approaches provide tamper-evident provenance and stronger Byzantine guarantees, but they target global participation and usually incur higher consensus cost~\cite{Aebeloe2021-WWW,zarina2021pbftlogd,yao2021convergenceEHR}. aFDOs address a different setting: small groups of partially trusted institutions that need auditable reconciliation of published assertions. Thus, the focus is not on where RDF is stored, but on how RDF assertions can evolve after publication.

\textbf{Agreement and scientific knowledge management.}
Byzantine fault tolerance has been studied since PBFT, including work on throughput, adversarial testing, and Linked Data publication~\cite{chen2022improvedpbft,winter2023byzzfuzz,zarina2021pbftlogd}. aFDOs do not provide a general-purpose BFT protocol. They use a reputation- and confidence-weighted aggregation mechanism for ordinal or numeric classifications and compare it with simpler voting alternatives in Section~\ref{sec:evaluation}. Related work in scientific knowledge management includes self-driving laboratories, contradiction detection with nanopublications, and FAIR Research Objects for reproducibility~\cite{bai2024dynamickgSDL,pavlo2017selfdriving,asif2021nanopubsContradictions,fouilloux2023fairROs}. The FAIR movement and FDO specifications support machine-actionable publication through persistent identifiers, metadata, and declared operations~\cite{wilkinson2016fair,wittenburg2022fair,lannom2022digital,schultes2022comparative}. aFDOs complement these efforts by adding policy, announcement, and agreement mechanisms to published knowledge artefacts~\cite{magagna2024nanopubsasfdo}.

aFDOs build on existing Web and FAIR standards rather than replacing them. Assertions use RDF-star, provenance records use PROV-O, identifiers use the Handle System or compatible URI schemes, operations follow DOIP, events use ActivityStreams 2.0, validation conditions use SHACL, and policy obligations use ODRL~\cite{rdfstar-cg,provo,lannom2022digital}. This mapping keeps policies, audit trails, validation conditions, and announcements portable across implementations.

\section{Autonomous FAIR Digital Objects}
\label{sec:afdo}

This section formalises aFDOs and presents the architecture. The model adds three components to a FAIR Digital Object: a policy specification that defines when actions fire, an event interface that connects observable events to policy evaluation, and a communication interface that exposes peer-facing operations and the agreement protocol. Together these address the three challenges from Section~\ref{sec:introduction}, namely continuous validation, discovery under growth, and cross-institution coordination. The remainder of this section gives the formal model, the architecture that realises it, and the RDF representation of each component, with empirical results deferred to Section~\ref{sec:evaluation}.

\subsection{Formal Model}
\label{sec:afdo-formal}

An aFDO extends a FAIR Digital Object with three components:
\begin{equation}
\text{aFDO} = \langle \text{FDO},\, \mathcal{P},\, \mathcal{E},\, \mathcal{C} \rangle
\end{equation}
where the underlying FDO $\langle \text{PID},\mathcal{T},\mathcal{O},\mathcal{M}\rangle$ provides identifier, type, declared operations, and metadata. The additional components are specifications rather than execution engines, and the implementation choices are discussed in Section~\ref{sec:evaluation}.

\paragraph{Policy specifications $\mathcal{P}$.}
A policy is a tuple $p = \langle \varphi,\, a,\, \kappa,\, \pi \rangle$ where $\varphi$ is a condition expressed as a SHACL shape or an N3 rule over RDF-star quoted triples~\cite{rdfstar-cg}, $a \in \mathcal{O} \cup \{\textsc{announce},\textsc{validate},\textsc{reconcile},\textsc{updateTrust}\}$ is the action, $\kappa$ is a set of ODRL-aligned obligations such as rate limits, temporal scope, and authorisation requirements, and $\pi$ is a PROV-O~\cite{provo} template for the audit record emitted on evaluation. Conditions read RDF-star-quoted triples to access claim-level metadata such as trust scores, source attribution, and confidence, which is the primary reason aFDOs use RDF-star rather than plain RDF.

\paragraph{Event interface $\mathcal{E}$.}
The event interface is a triple $\mathcal{E} = \langle \Sigma,\, \Phi,\, H \rangle$ where $\Sigma$ is an alphabet of event types drawn from ActivityStreams 2.0 activity names such as \texttt{Create}, \texttt{Update}, and \texttt{Announce}, $\Phi$ is a set of declarative subscription filters expressed as triple patterns over event payloads, and $H: \Sigma \to 2^{\mathcal{P}}$ binds each event type to the subset of policies whose evaluation it triggers.

\paragraph{Communication interface $\mathcal{C}$.}
The communication interface is a pair $\mathcal{C} = \langle \mathcal{O}_{\text{peer}},\, \Pi \rangle$ where $\mathcal{O}_{\text{peer}} \subseteq \mathcal{O}$ are the DOIP operations the aFDO exposes to peers, such as \texttt{seekClinicalValidation} or \texttt{negotiateClassification}, and $\Pi$ is the agreement protocol the aFDO participates in for multi-source updates, described in Section~\ref{sec:afdo-consensus}.

The triple $\langle \mathcal{P}, \mathcal{E}, \mathcal{C}\rangle$ is what makes the FDO active. $\mathcal{P}$ specifies \emph{when} an action may fire, $\mathcal{E}$ specifies \emph{which observable events} feed those decisions, and $\mathcal{C}$ specifies \emph{which peers and protocols} are reachable. Backward compatibility with FDO consumers is preserved, since a client that does not understand $\mathcal{P}$, $\mathcal{E}$, or $\mathcal{C}$ can still resolve the PID and call declared operations in $\mathcal{O}$.

\paragraph{Scope of decentralisation.}
aFDOs decentralise \emph{decision making}, not \emph{discovery}. This means decisions encoded in $\mathcal{P}$ are evaluated locally at the object layer, while discovery and event routing rely on a shared substrate of type and operation registries and an event bus, similar in role to DNS for the Web. Registry-free discovery through DHT-based mechanisms is discussed as future work in Section~\ref{sec:conclusion}. The prototype evaluated in this paper uses a registry-mediated substrate, and this dependency is treated as a limitation rather than as an eliminated component.

\subsection{Architecture}

\begin{figure}
    \centering
    \includegraphics[width=\linewidth]{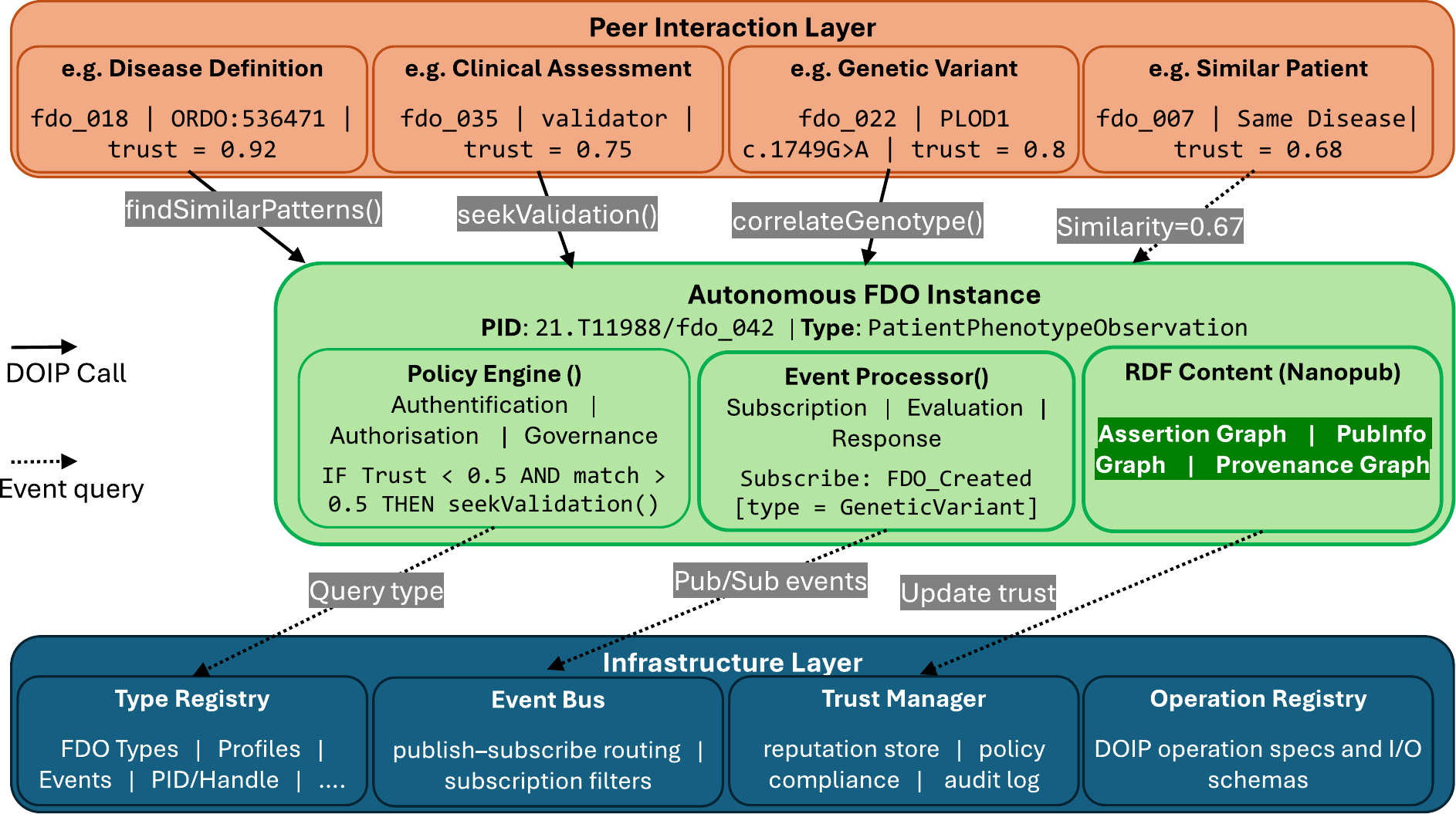}
    \caption{Three-layer aFDO architecture. The infrastructure layer provides registries, an event bus, and a trust register. The instance layer embeds policy specifications $\mathcal{P}$, the event interface $\mathcal{E}$, and FDO content. The peer interaction layer realises $\mathcal{C}$ via DOIP and event propagation. The figure shows a rare-disease instantiation, but the architecture is domain agnostic.}
    \label{fig:architecture}
\end{figure}

The architecture in Figure~\ref{fig:architecture} realises the formal model in three layers. The \textbf{infrastructure layer} provides shared services, namely type and operation registries that resolve $\mathcal{T}$ and $\mathcal{O}$, an event bus that routes elements of $\Sigma$ to subscribers matching $\Phi$, and a trust register that records reputation and audit entries. The \textbf{instance layer} embeds $\mathcal{P}$ and $\mathcal{E}$ within each aFDO together with its assertion, provenance, and pubinfo graphs. The \textbf{peer interaction layer} carries $\mathcal{C}$ through synchronous DOIP calls between aFDOs and asynchronous event propagation.

The announcement layer addresses discovery under growth without requiring exhaustive scans. On creation, an aFDO emits an announcement whose type belongs to $\Sigma$. The event bus matches it against subscription filters $\Phi$, and only peers whose filters match receive the announcement and run their SHACL condition $\varphi$. The work performed by any peer on a single announcement is bounded by the structure of its subscription, not by the size of the corpus. We do not validate the scaling behaviour of the announcement layer empirically, and leave systematic measurement of subscription-bounded discovery for future work.

\subsection{Continuous Validation Through Policy Specifications}
\label{sec:afdo-policy}

New evidence arrives asynchronously and may contradict existing claims, but manual triage does not scale. Each aFDO's policy specifications $\mathcal{P}$ define when validation should be sought and which peer operations may be invoked.

Policies are represented with SHACL conditions, ODRL obligations, and PROV-O audit templates over RDF-star assertions. For example, a patient observation aFDO may contain a policy that fires when its trust score is at most $0.5$, and its phenotype match score is at least $0.5$. The condition is expressed as an SHACL shape over the observation graph. The action invokes the DOIP operation \texttt{seekClinicalValidation}. An ODRL duty limits the action to one invocation per day. RDF-star is used to attach provenance to individual phenotype claims, rather than only to the nanopublication as a whole. The full Turtle serialisation is given in Appendix~\ref{app:policy-serialization}.

When the aFDO is created or its trust score changes, the SHACL condition is evaluated. If the condition holds and the ODRL duty permits execution, the action fires. The aFDO then invokes \texttt{seekClinicalValidation} on a clinical assessment peer reachable through $\mathcal{C}$. Every evaluation produces a PROV-O record in the audit log, regardless of whether the action fires.

\paragraph{Trust evolution.}
Validation outcomes and other observable events update the trust score:
\begin{equation}
T_{t+1} =
\begin{cases}
\min(1,\, T_t + \alpha)              & \text{validation confirmed} \\
T_t - \rho                           & \text{validation refuted} \\
T_t - \beta                          & \text{validation uncertain} \\
\min(1,\, T_t + \delta)              & \text{similar pattern found} \\
T_t - \beta \cdot \Delta\text{years} & \text{time since last update} \\
T_t \cdot (1 - \gamma)               & \text{institutional closure}
\end{cases}
\end{equation}
The coefficients are design parameters. The prototype uses $\alpha = 0.30$, $\beta = 0.05$, $\gamma = 0.20$, $\rho = 0.40$, and $\delta = 0.10$. Section~\ref{sec:eval-sensitivity} reports a sensitivity analysis for $\alpha$, $\beta$, and $\gamma$. Trust is decoupled from functionality: low-trust objects continue to operate and may recover through later confirming evidence.

\subsection{Cross-Institution Coordination Through Consensus}
\label{sec:afdo-consensus}

When two or more aFDOs assert contradictory interpretations of the same underlying object, identified by the same identifier in the assertion graph but with different classifications, the protocol $\Pi$ in $\mathcal{C}$ produces a consensus interpretation while preserving the original interpretations.

\paragraph{Conflict detection.}
For interpretations $V_A$ and $V_B$ over the same target identifier, conflict holds when $\text{ID}_A = \text{ID}_B$ and the asserted classifications differ. Detection is itself a policy. Each aFDO subscribes to announcements about its target and triggers the agreement protocol when the announced classification differs from its own.

\paragraph{Trimmed weighted-mean agreement.}
Classifications are mapped to scores $s \in [0,1]$ from a domain-specific scale. For ACMG variant interpretation, the mapping is Benign $= 0.0$, Likely Benign $= 0.25$, VUS $= 0.5$, Likely Pathogenic $= 0.75$, and Pathogenic $= 1.0$. Each contributing aFDO submits $\langle s_i, R_i, \text{conf}_i\rangle$, weighted by $w_i = R_i \cdot \text{conf}_i$. The protocol trims a fraction $\theta$ from each end of the ordered $s_i$ and computes the weighted mean of the remaining values, which is then mapped to a classification. The implementation uses $k = \max(1, \lfloor \theta \cdot n \rfloor)$ trimmed elements per side, which guarantees at least one element is removed from each end, even at small $n$.

The fraction $\theta$ controls the Byzantine threshold. Trimming $k$ elements from each end tolerates up to $k$ adversarial extremes per side, so $\theta = 0.20$ gives $f < n/5$ for $n \geq 5$. Section~\ref{sec:eval-byzantine} reports adversarial experiments with explicit Sybil, collusion, and evidence-poisoning attacks at varying $f$, and Section~\ref{sec:eval-sensitivity} varies $\theta$ over $\{0.05, 0.10, 0.15, 0.20, 0.25, 0.30, 0.40\}$. Section~\ref{sec:eval-byzantine} also analyses the choice between trimmed weighted-mean aggregation and simpler vote-counting alternatives, identifying tradeoffs that depend on the disagreement structure of the input.

\paragraph{Worked example.}
Consider five labs assessing the same PLOD1 variant, with reputations and confidences yielding weights $w = [0.15, 0.525, 0.60, 0.68, 0.765]$ and submitted scores $s = [0.0, 0.5, 0.75, 0.75, 1.0]$ in ascending order. With $\theta = 0.20$ and $n = 5$, the trim removes the extremes ($s = 0.0$ and $s = 1.0$). The weighted mean of the remaining three contributions is
\begin{equation*}
\frac{0.525 \cdot 0.5 + 0.60 \cdot 0.75 + 0.68 \cdot 0.75}{0.525 + 0.60 + 0.68} \approx 0.677,
\end{equation*}
which maps to Likely Pathogenic. The provenance record links all five inputs, the weights, and the policy version that selected $\theta$, which enables independent verification.
\section{Implementation and Evaluation}
\label{sec:evaluation}

We describe the reference implementation, the dataset and experimental setup, and five evaluation questions covering accuracy on real disagreements (\S\ref{sec:eval-realconf}), robustness under attack (\S\ref{sec:eval-byzantine}), parameter sensitivity (\S\ref{sec:eval-sensitivity}), the choice of aggregation strategy (\S\ref{sec:eval-ablation}), and correctness under distributed execution (\S\ref{sec:eval-wan}).

\subsection{Reference Implementation}
\label{sec:impl}

The reference implementation is a Python~3.10 system (\implLOC{} LOC) released under an open-source licence\footnote{\anonRepo{}}. Components correspond directly to the formal model. A \emph{policy module} compiles SHACL conditions against an in-memory RDF-star store and evaluates ODRL obligations. An \emph{event processor} subscribes to ActivityStreams 2.0 events and dispatches them to policies via the handler map $H$. A \emph{trust register} records reputation, audit entries, and PROV-O provenance. \emph{Registry adapters} resolve types, operations, and PIDs. A SHACL+ODRL \emph{serialiser} emits each policy in the form shown in Listing~\ref{lst:policy}, enabling round-trip validation through standard SHACL processors.

The implementation exposes seven DOIP operations, namely the four mandatory operations \texttt{Create}, \texttt{Retrieve}, \texttt{Update}, and \texttt{Delete}, together with \texttt{Search}, \texttt{ListOperations}, and a \texttt{Hello} extension required for capability discovery. On top of these, aFDOs declare four domain-level operations used in the experiments: \texttt{seekClinicalValidation}, \texttt{negotiateClassification}, \texttt{correlateWithGenotype}, and \texttt{findSimilarPatterns}. The event interface carries six ActivityStreams activity types: \texttt{Create}, \texttt{Update}, \texttt{Announce}, \texttt{Validate}, \texttt{Reconcile}, and \texttt{TrustChange}.

PIDs are assigned through the Handle System for the curated subset and through a containerised local Handle service for synthetic objects. The local service implements the Handle protocol's behavioural contract, which the rest of the stack uses, so PID resolution has identical semantics in both cases.

\subsection{Datasets}
\label{sec:datasets}

\paragraph{Rare-disease simulation corpus.} The simulation engine processes \nFDOs{} FDOs grounded in rare-disease ontologies. The corpus comprises 1{,}795 genetic variant interpretations from ClinVar, three disease definitions drawn from HPO and Orphanet (Ehlers-Danlos type VIA, hypermobile EDS, and kyphoscoliotic EDS), 1{,}509 synthetic patient observations grounded in real HPO terms, and 998 synthetic clinical assessments. Synthetic observations and assessments enable controlled injection of conflicts, institutional closures, and trust-evolution scenarios that are not externally observable in real data.

\paragraph{Real ClinVar contradictions.} For consensus evaluation we use \nRealConf{} variants for which ClinVar contains naturally occurring conflicting interpretations from at least two submitters with disagreement on pathogenicity classification. We filter to retain only those with a documented expert-panel resolution, which we use as ground truth held out from the consensus inputs. The dataset was built from the \clinvarRelease{} ClinVar release, with provenance and SHA-256 integrity recorded in the artefact. Submitter categories distribute as follows: clinical laboratories \cvCatA{}\%, individual submitters \cvCatB{}\%, research laboratories \cvCatC{}\%, with \nClinvarSubmissions{} non-expert-panel submissions across the \nRealConf{} records (mean 8.5 submissions per record).

\paragraph{Disagreement structure.} The ClinVar dataset partitions into four major-group disagreement buckets: \nClassA{} records where Pathogenic or Likely Pathogenic submissions disagree with VUS submissions, \nClassB{} records where VUS submissions disagree with Likely Benign or Benign submissions, \nClassC{} records of polar disagreement between Pathogenic/Likely Pathogenic and Likely Benign/Benign submissions, and \nClassD{} records where submissions span all three major groups simultaneously.

\paragraph{Adversarial scenarios.} We construct controlled adversarial conditions through three attack models: Sybil amplification, where one actor controls $f$ low-reputation submitters voting against the ground truth; reputation collusion, where the $f$ highest-reputation submitters coordinate on the same incorrect classification; and evidence poisoning, where $f$ randomly sampled submitters have their classifications replaced and their confidences boosted. The full sweep produces \nAdv{} consensus rounds across the Cartesian product of three attacks, seven adversary fractions $f/n \in \{0, 0.10, 0.15, 0.20, 0.25, 0.33, 0.50\}$, all \nRealConf{} records, and 50 trials per cell.

\subsection{Experimental Setup}
\label{sec:setup}

All experiments run on \hwSetup{}. For distributed experiments, we deploy a five-server stack across four geographically modelled regions (Europe, US-East, US-West, China) with synthetic WAN round-trip latency from cloudping.co. The mean cross-region RTT is \rttMean{} ms with standard deviation \rttStd{} ms. PYTHONHASHSEED is pinned to 0 across all runs, and the random seed is fixed at 42. Per-experiment random sequences derive from \texttt{numpy.random.SeedSequence(42).spawn()} hierarchically, which makes any cell of the sweep reproducible independently. Bootstrap confidence intervals use 1000 resamples.

\subsection{Real-Conflict Consensus Accuracy}
\label{sec:eval-realconf}

We test whether the trimmed weighted-mean consensus mechanism reproduces the expert-panel adjudication of \nRealConf{} ClinVar conflicts in which multiple submitters genuinely disagree. The expert-panel submission is removed from the consensus inputs and used only as ground truth. Consensus is computed from the remaining submissions, which average 8.5 per variant.

Overall accuracy is \realConfAcc{}\%. The results vary substantially across disagreement buckets (Table~\ref{tab:realconf}), reflecting the inherent difficulty of each case type. Polar disagreements between Pathogenic/Likely Pathogenic and Likely Benign/Benign submissions (\accClassC{}\%) are the easiest to resolve, since the evidence is concentrated at one extreme of the ordinal scale. Three-group spans, where submissions cover all three major groups, are the hardest (\accClassD{}\%), since no single class has a clear plurality.

We do not report accuracy against manual curation or random baselines. The former is a feasibility envelope rather than a quality comparison, and the latter is uninformative for a deterministic policy-driven system.

\begin{table}[t]
\centering
\caption{Consensus agreement with ClinVar expert-panel adjudications, stratified by disagreement bucket. Each bucket reports the number of records and the mean accuracy with 95\% bootstrap confidence interval.}
\label{tab:realconf}
\small
\begin{tabular}{lrr}
\toprule
\textbf{Disagreement bucket} & \textbf{N} & \textbf{Accuracy (\%)} \\
\midrule
Pathogenic/Likely Pathogenic vs.\ VUS  & \nClassA{} & \accClassA{} \\
VUS vs.\ Likely Benign/Benign           & \nClassB{} & \accClassB{} \\
Pathogenic/Likely Pathogenic vs.\ Likely Benign/Benign & \nClassC{} & \accClassC{} \\
All three major groups span             & \nClassD{} & \accClassD{} \\
\midrule
Overall                                  & \nRealConf{} & \realConfAcc{} \\
\bottomrule
\end{tabular}
\end{table}

\subsection{Adversarial Robustness}
\label{sec:eval-byzantine}

\begin{figure}[t]
\centering
\includegraphics[width=\linewidth]{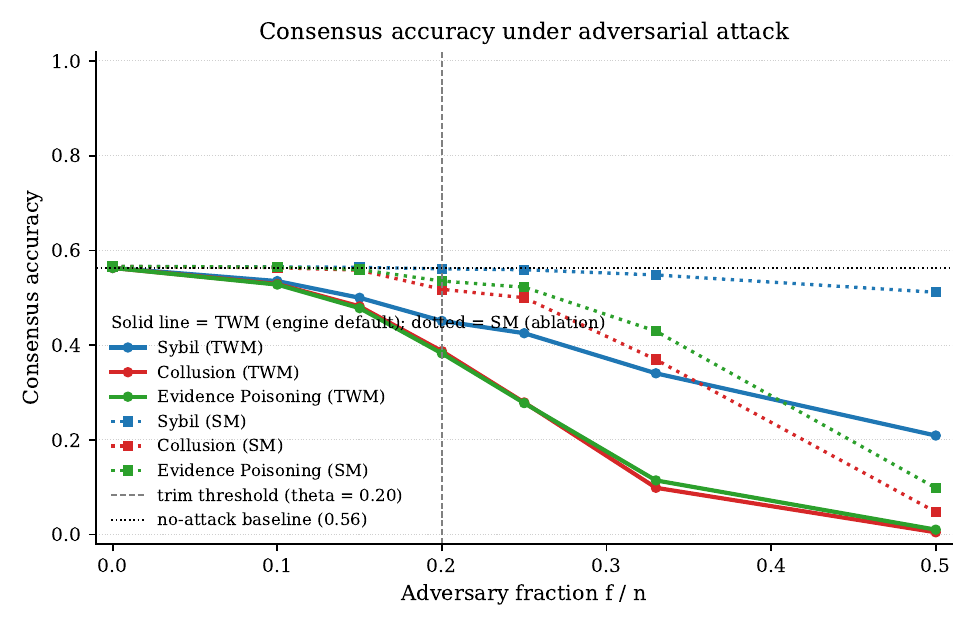}
\caption{Consensus accuracy under three attack models as the adversary fraction $f/n$ varies. Solid lines show trimmed weighted-mean (TWM, the engine default); dotted lines show simple majority (SM, the ablation analysed in \S\ref{sec:eval-ablation}). The vertical dashed line marks the trim threshold $\theta = 0.20$, which marks the design boundary.}
\label{fig:adv}
\end{figure}

Figure~\ref{fig:adv} reports consensus accuracy for trimmed weighted-mean (TWM, the engine default) under the three attack models, alongside simple majority (SM, included as an aggregation alternative analysed in \S\ref{sec:eval-ablation}) at each adversary fraction. The three TWM curves cluster tightly until the trim threshold $\theta = 0.20$, then diverge sharply.

Within the stated adversarial bound ($f < n/5$), TWM degrades gradually under all three attacks. At the boundary $f/n = 0.20$, Sybil causes the least damage, with TWM accuracy remaining above \advAccBound{}\%. Sybil submissions enter at low reputation, so the trim removes them before they affect the weighted mean. Beyond the design tolerance, the picture differs across attacks. At $f/n = 0.50$, TWM accuracy drops to 0.5\% under collusion and 1.1\% under poisoning, since both attacks corrupt the high-weight contributions that the trim preserves. Sybil retains 21\% even at $f/n = 0.50$ because low-reputation contributions remain at the score extremes that the trim removes.

This is a graceful-degradation result within the design tolerance and a predicted-failure result beyond it. The mechanism is principled against attackers with bounded reputation manipulation. The implemented attacks at $f \geq n/5$ are stronger than that bound and produce failures as predicted by the bound.

\subsection{Sensitivity to Parameters}
\label{sec:eval-sensitivity}

\paragraph{Trim ratio.}
We vary $\theta$ over the seven values $\{0.05, 0.10, 0.15, 0.20, 0.25, 0.30, 0.40\}$ on the baseline without attacks. Accuracy varies by less than $0.7$ percentage points across the entire range, with bootstrap confidence intervals overlapping at every point. The default $\theta = 0.20$ is $0.02$ percentage points below the empirical peak at $\theta = 0.30$, and is statistically indistinguishable from any other value tested. The mechanism is therefore robust to the choice of $\theta$ within this range. The corresponding plot is given in Appendix~\ref{app:eval-sensitivity}.

\paragraph{Trust dynamics.}
We vary the three trust evolution coefficients, $\alpha$, $\beta$, and $\gamma$, over an order of magnitude using a discrete-event simulator that exercises the trust update arithmetic in isolation. The main observations are stable. The coefficient $\alpha$ has the largest effect on recovery time after closure events. The coefficient $\beta$ shapes the final trust distribution. The coefficient $\gamma$ has the smallest effect on either metric. Within $\pm$\tsPerturbPct{}\% perturbations of each parameter at the default values, recovery time changes by \tA{}\% for $\alpha$, \tB{}\% for $\beta$, and \tG{}\% for $\gamma$. The Kolmogorov-Smirnov distance from the default trust distribution remains below \ksBound{} across all 27 grid cells. The full table is reported in Appendix~\ref{app:eval-sensitivity}.

\subsection{Consensus Strategy Ablation}
\label{sec:eval-ablation}

\begin{figure}[t]
\centering
\includegraphics[width=0.7\linewidth]{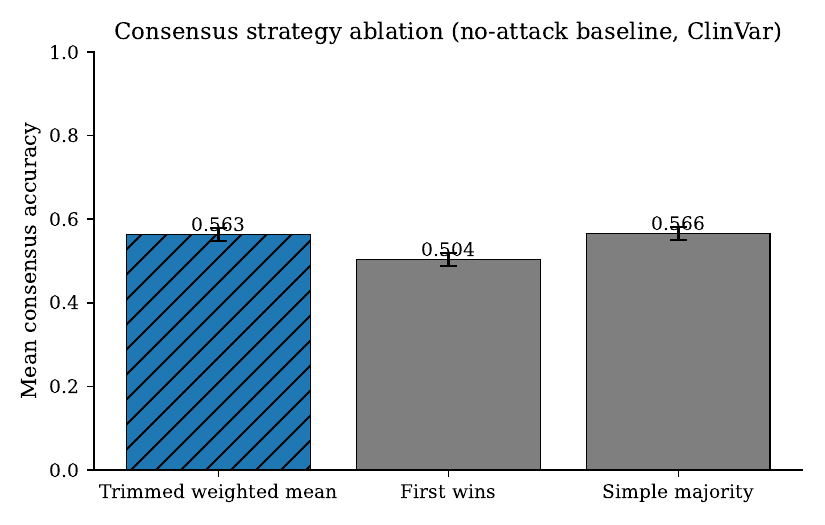}
\caption{Consensus accuracy across three aggregation strategies on the baseline without attacks. Trimmed weighted mean and simple majority are statistically tied, while first wins trails by about six percentage points.}
\label{fig:ablation}
\end{figure}

We compare the trimmed weighted mean strategy with two simpler alternatives: \emph{first wins}, where the earliest submitted classification is retained, and \emph{simple majority}, where a plurality vote is taken with equal weights. All three strategies are evaluated on the baseline without attacks over \nRealConf{} ClinVar conflicts (Figure~\ref{fig:ablation}).

First wins reaches \firstWinsAcc{}\% accuracy, trailing both alternatives by about six percentage points with confidence intervals that do not overlap. This confirms that aggregation contributes value over retaining the first submitted classification. Trimmed weighted mean reaches \realConfAcc{}\%, while simple majority reaches \majorityAcc{}\%. Their point estimates differ by less than $0.3$ percentage points, and their confidence intervals overlap.

Read together with the adversarial results in Figure~\ref{fig:adv}, this comparison shows a tradeoff. Overall, the trimmed weighted mean and simple majority are statistically tied on the baseline without attacks, with accuracies of \realConfAcc{}\% and \majorityAcc{}\%, respectively. A simple majority is often more robust to attacks because it ignores reputation and confidence, which are precisely the signals corrupted by collusion and poisoning. The trimmed weighted mean is more useful when the ordinal structure of the classification matters. On the polar disagreement bucket specifically, namely Pathogenic or Likely Pathogenic versus Likely Benign or Benign, it reaches \accClassC{}\% compared with \majorityAccClassC{}\% for simple majority. It also produces a continuous pathogenicity score that downstream applications can use. We therefore keep the trimmed weighted mean as the default, while treating the aggregation strategy as an application-specific design choice.

\subsection{Distributed Execution and State Equivalence}
\label{sec:eval-wan}

The system supports distributed FDO creation across the five servers, a four-region setup described in Section~\ref{sec:setup}. We compare three configurations on the same workload of 100 ClinVar conflict records: centralised execution in a single process, distributed execution without WAN latency, and distributed execution with synthetic WAN latency.

Across all 100 records and all three pairwise configuration comparisons, every FDO record is byte equivalent at the creation step, except for time-derived fields. The evolution layer, including trust dynamics, validation chains, contradiction resolution, similarity reinforcement, and time decay, runs as a centralised post-process in the current implementation. We identify the distribution of this layer as future work in Section~\ref{sec:conclusion}. The state equivalence result shows that the creation step does not require a single process, while the current evolution layer remains centralised.

Distributed execution adds a $\centLatRatio\times$ wall clock overhead compared to centralised execution. Distributed coordination accounts for most of the slowdown, while synthetic WAN latency adds an additional $12$\% on top of the coordination overhead. This suggests that improving the distributed coordination protocol would yield more benefit than reducing network latency. The full timing comparison is given in Appendix~\ref{app:eval-wan}.

\subsection{Engine Reproducibility}
\label{sec:eval-reproducibility}

The engine produces identical outputs across runs at fixed seed. We verify this through a two-run reproducibility check that compares simulation, evaluation, operation, and trust logs across independent invocations. All four log types are byte-equivalent across runs, as are the FDO records themselves. The check fixes \texttt{PYTHONHASHSEED=0} as a project-level invariant, sorts all globbed input file lists lexicographically, and disambiguates duplicate \texttt{prov:generatedAtTime} triples in input nanopublications by lexicographic minimum. These determinism fixes are documented in the artefact and apply to both the centralised and distributed execution paths.
\section{Conclusion}
\label{sec:conclusion}

In this paper, we addressed the gap between what FAIR Digital Objects publish and what active curation requires. We embedded condition-action policies, event handlers, and an agreement protocol within the artefact itself, formalised the result as Autonomous FAIR Digital Objects, and expressed the model in established W3C vocabularies rather than introducing a new representation. The reference implementation and its evaluation on real ClinVar disagreements show that policy-driven curation can be expressed in standards, evaluated against held-out expert ground truth, and deployed across geographically separated peers without losing state correctness. The ablation analysis revealed a finding that extends beyond this work: classification consensus on an ordinal scale faces a structural trade-off between weighted aggregation, which captures graded evidence at the cost of bin-boundary fragility under attack, and plurality voting, which is more robust to concentrated votes but cannot represent graded evidence. This tradeoff is not specific to ClinVar and applies to any consensus mechanism whose output is discretised from a continuous score. The broader contribution is methodological. The work shows that decision logic can live alongside the artefact and remain auditable through standard provenance vocabularies, which is a step toward ensuring FAIR data remains actionable after the institutions that publish it change or disappear.

The present implementation distributes only the FDO creation step. The trust dynamics, validation chain, and contradiction resolution run as a centralised post-process, and distributing them raises coordination questions that this paper does not address. Ground-truth-anchored evaluation is based on a single classification scale (ACMG variant interpretation), and generalisation to other ordinal tasks is a structural property of the architecture but is not empirically demonstrated here. The trust model assumes initial reputation is supplied externally, so cold-start from a fully decentralised state requires identity and reputation protocols beyond the scope of this paper. The discovery substrate uses a shared registry, and registry-free deployment via DHT-based routing remains future work. Three directions follow from these limitations: distributing the evolution layer to convert the centralised post-process into peer-coordinated trust dynamics, integrating with existing nanopublication repositories to operate over the deployed nanopub corpus, and formal analysis of policy composition to put the autonomous behaviour on firmer footing than the empirical evidence presented here.


\section*{Declaration of use of Generative AI}
The authors used Generative AI tools to support the preparation of the manuscript, including language refinement, grammar checking, clarity improvements in the presentation, and code development. All AI-assisted outputs were reviewed, verified, and edited by the authors. The authors remain fully responsible for the content, claims, implementation, results, and conclusions of the paper.

\paragraph*{Supplemental Material Statement:}
The source code required to reproduce the experiments reported in this paper is available at \anonRepo. Additional supplementary material, including extended details and results supporting reproducibility and verification, is provided in the supplementary material and the artefact.

\bibliographystyle{splncs04}
\bibliography{references}

\appendix
\appendix

\section*{Appendix: Supplementary Material}
\label{sec:app}

This appendix provides supplementary material supporting the claims in the main paper. It includes policy serialisation examples (\S\ref{app:policy-serialization}), the design intent and empirical evidence for the bounded adversarial model (\S\ref{app:byzantine}), the construction of the real-conflict dataset (\S\ref{app:clinvar}), the adversarial attack models (\S\ref{app:adversarial}), extended SHACL and ODRL examples (\S\ref{app:standards}), parameter sensitivity sweeps (\S\ref{app:sensitivity}), implementation details (\S\ref{app:impl}), and ethics and governance considerations (\S\ref{app:ethics}).

\section{Policy Serialisation Example}
\label{app:policy-serialization}

Listing~\ref{lst:policy} gives the Turtle serialisation of the policy example from Section~\ref{sec:afdo-policy}. The policy uses a SHACL condition, a DOIP action, and an ODRL duty. RDF-star is used to attach provenance to an individual phenotype assertion.

\begin{lstlisting}[caption={Patient observation aFDO with SHACL and ODRL policy in Turtle.},label={lst:policy},basicstyle=\ttfamily\scriptsize,frame=single]
@prefix afdo: <http://w3id.org/afdo#> .
@prefix prov: <http://www.w3.org/ns/prov#> .
@prefix sh:   <http://www.w3.org/ns/shacl#> .
@prefix odrl: <http://www.w3.org/ns/odrl/2/> .
@prefix xsd:  <http://www.w3.org/2001/XMLSchema#> .
@prefix hp:   <http://purl.obolibrary.org/obo/HP_> .

:obs042 a afdo:PatientPhenotypeObservation ;
        afdo:hasPhenotype hp:0001382, hp:0000974 ;
        afdo:trustScore 0.45 ;
        afdo:phenotypeMatchScore 0.72 .

<< :obs042 afdo:hasPhenotype hp:0001382 >>
        prov:wasDerivedFrom :clinicalRecord_X .

:obs042-policy a afdo:Policy ;
    afdo:condition [
        a sh:NodeShape ;
        sh:targetNode :obs042 ;
        sh:property [
            sh:path afdo:trustScore ;
            sh:maxInclusive 0.5 ] ;
        sh:property [
            sh:path afdo:phenotypeMatchScore ;
            sh:minInclusive 0.5 ] ] ;
    afdo:action afdo:seekClinicalValidation ;
    odrl:duty [
        a odrl:Duty ;
        odrl:action odrl:rateLimit ;
        odrl:constraint [
            odrl:leftOperand odrl:elapsedTime ;
            odrl:operator odrl:gteq ;
            odrl:rightOperand "P1D"^^xsd:duration ] ] .
\end{lstlisting}

\section{Byzantine Properties}
\label{app:byzantine}

The trimmed weighted-mean consensus mechanism specified in \S\ref{sec:afdo-consensus} is designed to provide specific guarantees within the design Byzantine tolerance $f < n/5$. We state four properties below as design intent, grounded in the mechanism's structure and supported empirically by the adversarial sweep in \S\ref{sec:eval-byzantine}. Formal proofs are left to future work, as noted in \S\ref{sec:conclusion}.

\paragraph{Safety.} If all honest participants propose values in range $[v_{\min}, v_{\max}]$, and the number of Byzantine participants $f$ satisfies $f < n/5$, the consensus output $v^*$ satisfies $v_{\min} \leq v^* \leq v_{\max}$. The trim of $k = \max(1, \lfloor \theta \cdot n \rfloor)$ values from each end of the ordered submissions removes Byzantine extremes within the trimmed region, leaving a weighted mean computed over honest contributions. Beyond $f < n/5$ this property does not hold, as Byzantine submissions can survive the trim through collusion or poisoning. The empirical results in \S\ref{sec:eval-byzantine} show this transition clearly.

\paragraph{Liveness.} Under partial synchrony, where messages are eventually delivered within an unknown bound, consensus terminates in a single round when at least $n - f$ honest participants remain responsive. The protocol does not require multiple voting rounds, as the trimmed weighted mean is computed directly from the received submissions.

\paragraph{Agreement.} All honest participants that complete the protocol compute the same consensus value $v^*$ up to floating-point precision. Determinism follows from the trim and weighted-mean computations being functions only of the sorted weighted submission set, which is identical for all honest participants observing the same set of submissions.

\paragraph{Validity.} The consensus value $v^*$ is a weighted combination of at least $n - 2k$ submissions after trimming $k$ from each end. With $\theta = 0.20$ and $n \geq 5$, this leaves at least three contributors in the weighted mean.

\paragraph{Synchrony assumption.} Partial synchrony assumes there exists an unknown global stabilisation time GST after which all messages between honest participants are delivered within a known bound $\Delta$. Before GST, network behaviour is arbitrary, including partitions. The implementation enforces a per-round timeout of \consTimeout{} seconds. Submissions arriving after the timeout are excluded from the round but contribute to subsequent rounds for the same target.

\paragraph{Identity assumption.} The protocol assumes participants have stable identifiers that persist across consensus rounds. The reference implementation realises this through Handle System PIDs for institutional aFDOs, and through public-key signatures over submissions for synthetic participants in adversarial experiments. We do not contribute a Sybil-resistant identity layer. Cold-start identity bootstrapping in a fully decentralised setting remains future work.

\section{ClinVar Conflict Dataset Construction}
\label{app:clinvar}

The real-conflict experiment in \S\ref{sec:eval-realconf} draws ground-truth disagreements from ClinVar~\cite{landrum2018clinvar}, the NCBI public archive of genetic variant interpretations. This appendix describes the dataset construction pipeline.

\paragraph{Source data.} We use ClinVar release \clinvarRelease{}, accessed through the NCBI public FTP at \clinvarURL{}. The release contains \nClinvarVariants{} unique variants. The dataset identity is pinned by the SHA-256 of the source bulk files rather than by URL, since NCBI rotates the contents of the URL between releases without preserving historical snapshots. The release.json file in the artefact records the exact bytes used.

\paragraph{Conflict identification.} A variant is included in the conflict dataset if it satisfies all of the following criteria. First, at least two distinct submitters have submitted classifications for it. Second, after joining per-submitter records, the submissions span at least two of the three major pathogenicity groups: Pathogenic/Likely Pathogenic, VUS, and Benign/Likely Benign. Third, at least one submission carries an expert-panel review status (\texttt{reviewed by expert panel} or \texttt{practice guideline}), which we treat as ground truth and remove from the consensus inputs. Fourth, the variant is not flagged as lacking assertion criteria. ClinVar removes withdrawn variants from the bulk file between releases, so any variant present in the release with non-empty classification data is implicitly current. The third criterion is the binding one. It restricts the dataset to variants for which a defensible ground truth exists, at the cost of excluding variants where disagreement persists in the literature without authoritative resolution.

\paragraph{Statistics.} Applying the filter yields \nRealConf{} variants with both submitter disagreement and expert-panel adjudication. Table~\ref{tab:clinvar-construction} reports the counts at each filter stage.

\begin{table}[h]
\centering
\caption{ClinVar conflict dataset filter pipeline.}
\label{tab:clinvar-construction}
\small
\begin{tabular}{lr}
\toprule
\textbf{Stage} & \textbf{Variants} \\
\midrule
Total variants in release \clinvarRelease{} & \nClinvarVariants{} \\
With $\geq 2$ distinct submitters         & \cvFiltA{} \\
With expert-panel adjudication present     & 36{,}830 \\
With $\geq 2$ non-expert-panel submissions & \cvFiltB{} \\
With major-class disagreement (final)      & \cvFiltC{} \\
\bottomrule
\end{tabular}
\end{table}

\paragraph{Submitter reputation initialisation.} For each submitter contributing to the conflict set, an initial reputation $R$ is assigned based on the submitter category derived from the submitter name. Clinical laboratories receive $R = 0.85$, research laboratories receive $R = 0.70$, and individual or unattributed submitters receive $R = 0.55$. Expert-panel submissions are treated separately as ground truth and not included in the consensus inputs. The submitter category breakdown across the \nClinvarSubmissions{} non-expert-panel submissions is clinical laboratories \cvCatA{}\%, individual submitters \cvCatB{}\%, and research laboratories \cvCatC{}\%.

\paragraph{Confidence assignment.} Submission confidence is derived from ClinVar's review-status field on a per-submission basis. The mapping is \texttt{criteria provided, multiple submitters, no conflicts} to $0.85$, \texttt{criteria provided, single submitter} to $0.70$, \texttt{criteria provided, conflicting interpretations} to $0.55$, and \texttt{no assertion criteria provided} to $0.40$.

\paragraph{Submitter hashing.} Submitter names are hashed to eight-character hexadecimal identifiers using SHA-256 with a fixed project salt. The same submitter name produces the same hash across runs, but the hash cannot be reversed to recover the name. The salt and hashing function are documented in the artefact, so reviewers can verify hashes given the original ClinVar submitter names.

\paragraph{Example record.} Listing~\ref{lst:clinvar-record} shows one entry from the conflict dataset in the internal JSON format.

\begin{lstlisting}[caption={ClinVar conflict record (abbreviated, JSON).},label={lst:clinvar-record},basicstyle=\ttfamily\scriptsize,frame=single]
{
  "variant_id": "VCV000017896",
  "hgvs": "NM_000169.3:c.1234G>A",
  "gene": "GLA",
  "ground_truth": {
    "submitter_hash": "h_panel_abc12345",
    "category": "expert_panel",
    "classification": "Pathogenic",
    "review_status": "reviewed_by_expert_panel"
  },
  "submissions": [
    {"submitter_hash": "h1a2b3c4", "category": "clinical_lab",
     "classification": "Pathogenic",
     "review_status": "criteria_provided_single_submitter",
     "R": 0.85, "conf": 0.70},
    {"submitter_hash": "h5d6e7f8", "category": "research_lab",
     "classification": "VUS",
     "review_status": "criteria_provided_single_submitter",
     "R": 0.70, "conf": 0.70}
  ]
}
\end{lstlisting}

The release artefact includes the construction script, the version-pinned ClinVar dump used as input, and the resulting conflict set.

\section{Adversarial Attack Models}
\label{app:adversarial}

The adversarial sweep in \S\ref{sec:eval-byzantine} evaluates three attack models. This appendix specifies each.

\paragraph{Sybil amplification.} An adversary controls multiple low-reputation submitters that vote in concert against a target classification. Given a target variant with true classification $c^*$, the attacker creates $f$ Sybil submitters with reputation drawn uniformly from $[0.20, 0.40]$ and confidence drawn uniformly from $[0.40, 0.70]$, all submitting the same incorrect classification $c^- \neq c^*$. Honest submitters retain their original reputations and confidence from the ClinVar-derived dataset.

\begin{algorithm}[h]
\caption{Sybil amplification attack}
\begin{algorithmic}
\State \textbf{Input:} variant $v$, true class $c^*$, attack count $f$
\State $H \gets$ honest submissions for $v$
\State $S \gets \emptyset$
\For{$i = 1$ to $f$}
    \State $R_i \sim \mathcal{U}(0.20, 0.40)$
    \State $\text{conf}_i \sim \mathcal{U}(0.40, 0.70)$
    \State $c_i \gets c^-$ \Comment{Adversarial alternative to $c^*$}
    \State $S \gets S \cup \{\langle c_i, R_i, \text{conf}_i\rangle\}$
\EndFor
\State \Return $H \cup S$
\end{algorithmic}
\end{algorithm}

\paragraph{Reputation collusion.} A subset of high-reputation submitters coordinates on the same incorrect classification. This is a stronger attack than Sybil, because the colluding submissions are not extreme values that the trim removes. Their high reputation positions them toward the centre of the weighted distribution, thereby directly influencing the weighted mean. The attacker selects $f$ submitters with the highest reputations from the honest set and reassigns their classifications to $c^-$ while keeping reputations and confidences unchanged.

\paragraph{Evidence poisoning.} A submitter contributes a semantically well-formed but factually wrong submission with high confidence. The attacker samples $f$ submitters from the full honest set, sets their confidence to the maximum observed in the round, and reassigns their classification to $c^-$. Earlier versions of this attack restricted sampling to the top reputation quartile, producing an artefactual saturation in the accuracy curve at high $f/n$ when the quartile was exhausted; the version reported here samples from the full set, yielding a continuous degradation curve.

For each attack model and each sweep value $f/n \in \{0, 0.10, 0.15, 0.20, 0.25, 0.33, 0.50\}$, we run \nAdvRuns{} trials with different attack-target selections and report mean accuracy with 95\% bootstrap confidence intervals in Figure~\ref{fig:adv}.

\section{Extended SHACL+ODRL Examples}
\label{app:standards}

The main paper shows one Turtle policy in Listing~\ref{lst:policy}. Listing~\ref{lst:consensus-policy} shows the consensus-trigger policy used by genetic variant interpretation aFDOs.

\begin{lstlisting}[caption={Consensus-trigger policy (Turtle).},label={lst:consensus-policy},basicstyle=\ttfamily\scriptsize,frame=single]
:variant-policy a afdo:Policy ;
    afdo:condition [
        a sh:NodeShape ;
        sh:targetClass afdo:GeneticVariantInterpretation ;
        sh:property [
            sh:path afdo:variantId ;
            sh:equals [ sh:path :announced/afdo:variantId ] ] ;
        sh:property [
            sh:path afdo:classification ;
            sh:not [ sh:equals
                [ sh:path :announced/afdo:classification ] ] ]
    ] ;
    afdo:action  afdo:negotiateClassification ;
    odrl:duty    [ a odrl:Duty ;
                   odrl:action  odrl:notify ;
                   odrl:assignee afdo:TrustRegister ] .
\end{lstlisting}

The condition uses SHACL property paths to compare the receiving aFDO's variant identifier and classification with those carried in the announcement payload. When the variant identifiers match but the classifications differ, the action invokes \texttt{negotiateClassification}, and the ODRL duty requires that the trust register be notified for audit purposes.

\paragraph{Round-trip serialisation.} The reference implementation includes a serialiser that emits each policy in this Turtle form and a parser that reads SHACL+ODRL Turtle into the internal policy representation. We verified round-trip equivalence on the policy set used in the experiments. Every internal policy was serialised to Turtle, validated against a SHACL meta-shape using a standard validator, parsed back into the internal representation, and checked for behavioural equivalence on a fixed test set of \nRoundTrip{} evaluation inputs. All policies passed. The serialiser, meta-shape, and validation script are included in the artefact.

\section{Full Sensitivity Sweeps}
\label{app:sensitivity}

The main paper reports summary statistics from the parameter sensitivity analysis in \S\ref{sec:eval-sensitivity}. This appendix provides the full sweep tables.

\paragraph{Trim ratio $\theta$.} Table~\ref{tab:full-trim} reports overall consensus accuracy on the ClinVar conflict dataset across $\theta \in \{0.05, 0.10, 0.15, 0.20, 0.25, 0.30, 0.40\}$, alongside per-bucket accuracy at each value. The chosen $\theta = 0.20$ corresponds to the design Byzantine bound $f < n/5$ for $n \geq 5$.

\begin{table}[h]
\centering
\caption{Consensus accuracy as a function of trim ratio $\theta$, overall and per disagreement bucket. All values in per cent.}
\label{tab:full-trim}
\small
\begin{tabular}{lrrrrrrr}
\toprule
\textbf{$\theta$} & 0.05 & 0.10 & 0.15 & 0.20 & 0.25 & 0.30 & 0.40 \\
\midrule
P/LP vs.\ VUS                  & \trAa{} & \trAb{} & \trAc{} & \trAd{} & \trAe{} & \trAf{} & \trAg{} \\
VUS vs.\ LB/B                   & \trBa{} & \trBb{} & \trBc{} & \trBd{} & \trBe{} & \trBf{} & \trBg{} \\
P/LP vs.\ LB/B                  & \trCa{} & \trCb{} & \trCc{} & \trCd{} & \trCe{} & \trCf{} & \trCg{} \\
Three-group span                & \trDa{} & \trDb{} & \trDc{} & \trDd{} & \trDe{} & \trDf{} & \trDg{} \\
\midrule
Overall                         & \trOa{} & \trOb{} & \trOc{} & \trOd{} & \trOe{} & \trOf{} & \trOg{} \\
\bottomrule
\end{tabular}
\end{table}

\begin{figure}[h]
\centering
\includegraphics[width=\linewidth]{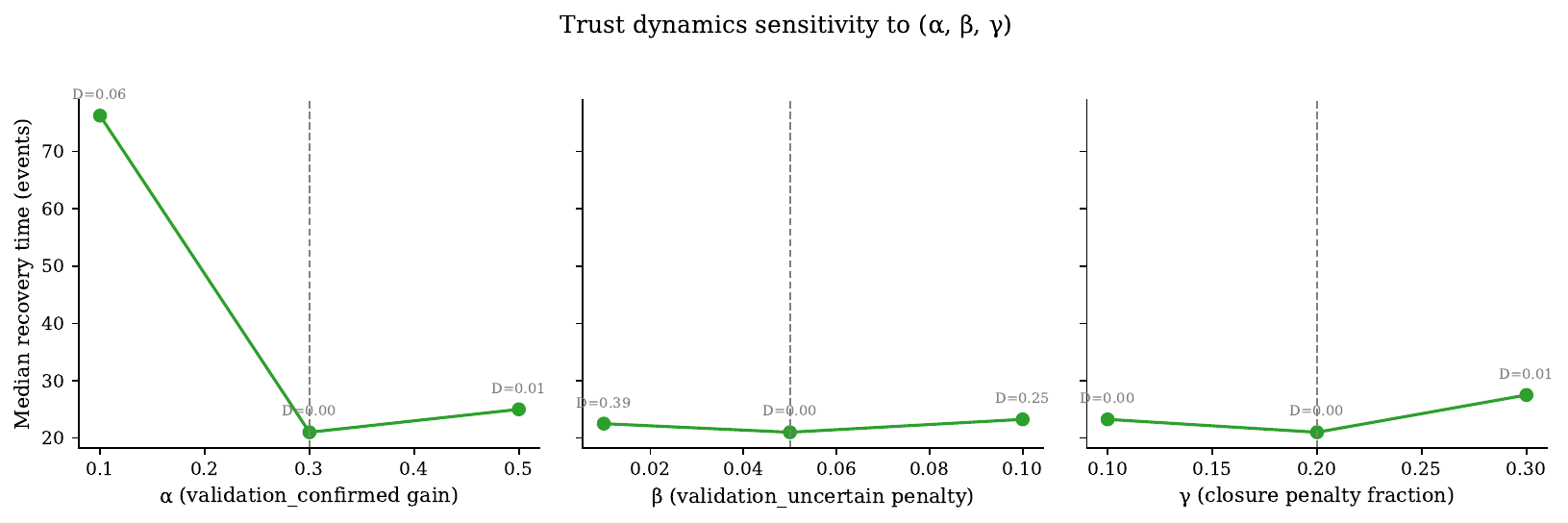}
\caption{Trust dynamics sensitivity to $(\alpha, \beta, \gamma)$. Each panel fixes two parameters at their defaults and varies the third. The vertical dashed line marks the default value. Annotations report the Kolmogorov-Smirnov distance from the default trust distribution.}
\label{fig:trust-sens}
\end{figure}

\paragraph{Trust dynamics $(\alpha, \beta, \gamma)$.} Table~\ref{tab:full-trust} reports median trust-recovery time after simulated institutional closure for the parameter cross-product. Each cell averages over 30 replicates with different closure timings. Figure~\ref{fig:trust-sens} presents the same data graphically, with one panel per parameter and KS distance from the default distribution annotated.

\begin{table}[h]
\centering
\caption{Median trust-recovery time (validation events) across $(\alpha, \beta, \gamma)$. Default in bold.}
\label{tab:full-trust}
\small
\begin{tabular}{ll|rrr}
\toprule
$\alpha$ & $\beta$ & $\gamma=0.10$ & $\gamma=0.20$ & $\gamma=0.30$ \\
\midrule
0.10 & 0.05 & \tsAa{} & \tsAb{} & \tsAc{} \\
0.30 & 0.05 & \tsBa{} & \textbf{\tsBb{}} & \tsBc{} \\
0.50 & 0.05 & \tsCa{} & \tsCb{} & \tsCc{} \\
\midrule
0.30 & 0.01 & \tsDa{} & \tsDb{} & \tsDc{} \\
0.30 & 0.10 & \tsEa{} & \tsEb{} & \tsEc{} \\
\bottomrule
\end{tabular}
\end{table}

The chosen defaults $(\alpha, \beta, \gamma) = (0.30, 0.05, 0.20)$ sit in a stable region. The minimum recovery time across the grid occurs at the default $\alpha = 0.30$, recovery time more than triples when $\alpha$ drops to $0.10$, and the final trust distribution shifts most strongly with $\beta$.

\paragraph{Trust dynamics simulator.} The trust dynamics sweep is run with a discrete-event simulator that exercises the trust-update arithmetic from the main engine in isolation, because the full simulation engine generates trust events at a rate that would make a 27-cell sensitivity sweep impractical. The arithmetic is byte-equivalent to the engine's \texttt{execute\_updateConfidence} function, verified through a paired test on identical input event sequences. The engine itself exercises the trust rules end-to-end, but at a single parameter setting per run.

\section{Implementation Details}
\label{app:impl}

\paragraph{System components.} The reference implementation is a Python 3.10 system using RDFLib for nanopublication parsing, pyshacl for SHACL evaluation, and a custom ODRL evaluator. The registry hierarchy includes four FDO type definitions, four metadata schemas anchored in HPO, ORDO, and HGVS terms, seven DOIP operation specifications with input and output JSON schemas, policy templates, and six ActivityStreams event definitions. All registry content is defined in JSON and published with the artefact.

\paragraph{DOIP communication.} Inter-aFDO communication uses DOIP over HTTPS. Each aFDO instance runs a small server that exposes its declared operations and listens for events from the bus. The Handle System adapter retrieves FDO records by PID, and for synthetic objects in scaling experiments, the local Handle service implements the same DOIP behavioural contract.

\paragraph{Logging.} Logs fall into four categories. The first is FDO record snapshots in JSON, including PID, type, current state, trust score, and operation history. The second is PM4Py-compatible XES event logs for process mining. The third is a debug trace covering every policy evaluation, event delivery, and consensus round. The fourth is a Turtle dump of all aFDO content for offline RDF analysis.

\paragraph{Determinism invariants.} The engine fixes \texttt{PYTHONHASHSEED=0} as a project-level invariant, sorts all globbed input file lists lexicographically, and disambiguates duplicate \texttt{prov:generatedAtTime} triples in input nanopublications by lexicographic minimum. Two independent runs with the same random seed produce byte-for-byte simulation, evaluation, operation, and trust logs. The reproducibility check script in the artefact runs the engine twice into separate output directories and compares all outputs at the byte level.

\paragraph{Resource consumption.} Across all experiments the system uses approximately \resMem{} peak memory and \resCPU{}.

\section{Ethics and Governance}
\label{app:ethics}

\paragraph{Data protection.} All patient phenotype observations and clinical assessments used in the experiments are synthetic. The only real personal-data-adjacent inputs are HPO and Orphanet ontology terms, which are publicly licensed. No HIPAA or GDPR-protected health information is processed. Real-world clinical deployment of aFDOs would require additional controls beyond the technical scope of this paper, including de-identification protocols, consent mechanisms for autonomous data sharing, and data sovereignty controls over which institutions may host an FDO.

\paragraph{Clinical decision boundaries.} Autonomous validation provides decision support, not clinical diagnosis. Confidence thresholds are set conservatively, with the default $0.5$ for triggering validation seek, and clinical-assessment outputs include a research-only disclaimer. The system is not intended for, and should not be deployed in, settings where treatment selection or diagnosis is conducted without human review.

\paragraph{Audit and traceability.} Every policy evaluation and consensus round produces a PROV-O record retained in the trust register. Audit trails support the integrity, time-stamping, and electronic-record requirements of FDA 21 CFR Part 11, provide a rationale for GDPR Article 22 automated-decision explanation, and are compatible with ISO 15189 traceability requirements for medical laboratories. Logs are append-only at the policy layer.

\paragraph{Bias and fairness.} Autonomous decisions can propagate biases present in their training and reference data. Rare-disease ontology terms reflect the populations from which the underlying clinical literature is drawn, which is known to skew toward European populations. Mitigation requires periodic bias audits comparing decision patterns across demographic groups, diversity considerations in trust assignment, and an open appeals process. We do not contribute a formal fairness evaluation in this work.

\paragraph{Liability and accountability.} Legal accountability for autonomous decisions is unsettled. The model in this paper assigns primary liability for content to the publishing institution (the FDO owner), bounds autonomous behaviour through operator-set policies, and uses audit logs as evidence for downstream liability determination. Transparency takes priority over full automation. Any decision can be traced back to the policy version that authorised it.

\paragraph{Appropriate use.} aFDOs as described here are appropriate for research knowledge management, non-critical decision support, institutional knowledge preservation, and collaborative scientific curation. They are not appropriate for high-stakes clinical decision-making without human review, real-time safety-critical systems, adversarial environments without strengthened identity guarantees beyond those evaluated in \S\ref{app:adversarial}, or proprietary or confidential knowledge that cannot be exposed in audit logs.

\section{Additional Evaluation Details}
\label{app:additional-evaluation}

\subsection{Parameter Sensitivity}
\label{app:eval-sensitivity}

\begin{figure}[t]
\centering
\includegraphics[width=\linewidth]{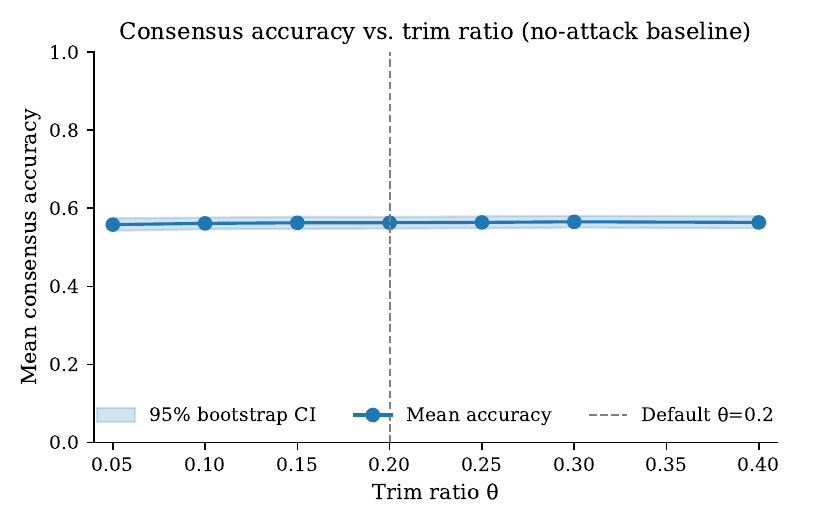}
\caption{Consensus accuracy as a function of the trim ratio $\theta$, with 95\% bootstrap confidence intervals. The dashed vertical line marks the default $\theta = 0.20$. Accuracy is statistically indistinguishable across the tested range.}
\label{fig:trim}
\end{figure}

Figure~\ref{fig:trim} reports consensus accuracy as a function of the trim ratio $\theta$. Accuracy varies by less than $0.7$ percentage points across the tested range, and all bootstrap confidence intervals overlap.

Table~\ref{tab:sens-trust} reports the sensitivity of trust dynamics to $\alpha$, $\beta$, and $\gamma$ around their default values.

\begin{table}[t]
\centering
\caption{Sensitivity of trust dynamics to $\alpha$, $\beta$, and $\gamma$ at $\pm$\tsPerturbPct{}\% perturbation around the default values $\alpha = 0.30$, $\beta = 0.05$, and $\gamma = 0.20$.}
\label{tab:sens-trust}
\small
\begin{tabular}{lrrr}
\toprule
\textbf{Parameter} & \textbf{Range tested} & \textbf{Recovery $\Delta$ (\%)} & \textbf{KS $D$ vs.\ default} \\
\midrule
$\alpha$ (validation gain)    & $0.10$--$0.50$ & \tA{} & 0.06 \\
$\beta$ (uncertainty penalty) & $0.01$--$0.10$ & \tB{} & 0.39 \\
$\gamma$ (closure penalty)    & $0.10$--$0.30$ & \tG{} & 0.01 \\
\bottomrule
\end{tabular}
\end{table}

\subsection{Distributed Execution}
\label{app:eval-wan}

\begin{figure}[t]
\centering
\includegraphics[width=\linewidth]{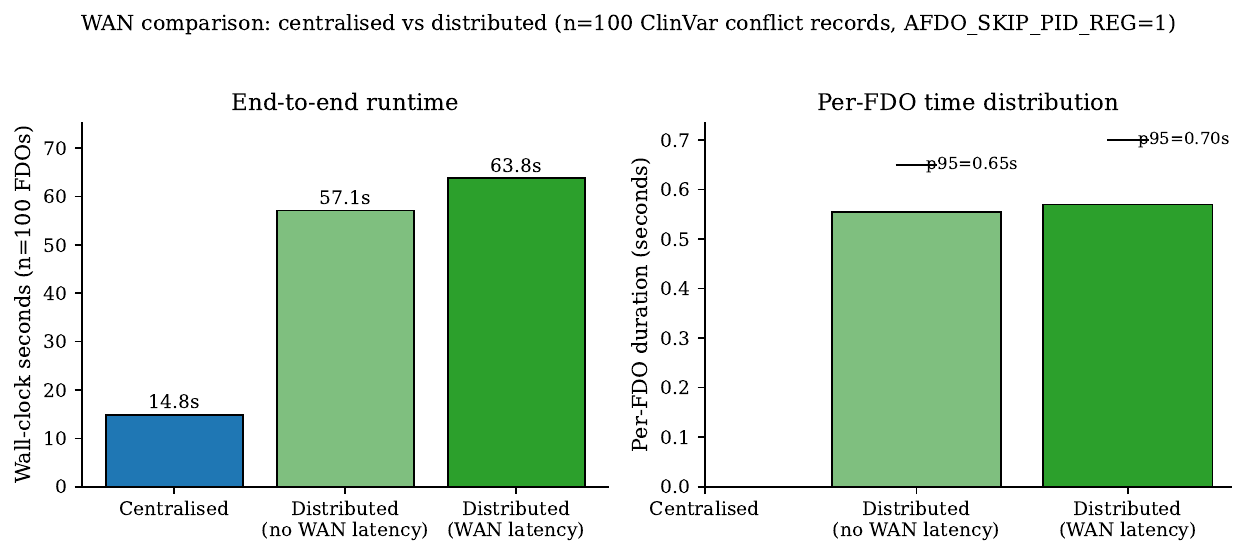}
\caption{WAN comparison on 100 ClinVar conflict records. Left: wall clock time across centralised, distributed without WAN latency, and distributed with WAN latency configurations. Right: per FDO time distribution for the two distributed configurations. Handle.Net registration is disabled in all modes to isolate consensus and coordination overhead from external service latency.}
\label{fig:wan}
\end{figure}

Figure~\ref{fig:wan} reports the timing comparison for centralised and distributed execution. Distributed coordination accounts for a $3.86\times$ slowdown, as seen in the no-latency configuration, which takes $57.1$ seconds compared with $14.8$ seconds for centralised execution. WAN latency adds an additional $12$\% on top of coordination overhead, increasing runtime from $57.1$ seconds to $63.8$ seconds. Per FDO p95 latency is \autoPNF{} seconds in distributed mode, compared with a centralised throughput of \centThru{} FDOs per second.

\end{document}